\def\year{2020}\relax
\documentclass[letterpaper]{article} % DO NOT CHANGE THIS
\usepackage{aaai20}  % DO NOT CHANGE THIS
\usepackage{times}  % DO NOT CHANGE THIS
\usepackage{helvet} % DO NOT CHANGE THIS
\usepackage{courier}  % DO NOT CHANGE THIS
\usepackage[hyphens]{url}  % DO NOT CHANGE THIS
\usepackage{graphicx} % DO NOT CHANGE THIS
\usepackage{graphicx}  % DO NOT CHANGE THIS
\usepackage{amsmath}
\usepackage{booktabs}
\usepackage{multirow}
\usepackage{textcomp}
\usepackage{graphics}
\usepackage{subcaption}

\title{Online Knowledge Distillation with Diverse Peers}
\author{Defang Chen,\textsuperscript{\rm 1,2} Jian-Ping Mei\textsuperscript{\rm 3}\thanks{Corresponding author}\\ \Large \textbf{Can Wang,\textsuperscript{\rm 1,2} Yan Feng,\textsuperscript{\rm 1,2} Chun Chen\textsuperscript{\rm 1,2}}\\ % All authors must be in the same font size and format. Use \Large and \textbf to achieve this result when breaking a line
\textsuperscript{\rm 1}College of Computer Science, Zhejiang University, Hang Zhou, China.\\ %If you have multiple authors and multiple affiliations
\textsuperscript{\rm 2}ZJU-LianlianPay Joint Research Center. \\
\textsuperscript{\rm 3}College of Computer Science, Zhejiang University of Technology, Hang Zhou, China. \\
defchern@zju.edu.cn, jpmei@zjut.edu.cn, \{wcan, fengyan, chenc\}@zju.edu.cn
}

\begin{document}

\maketitle

\begin{abstract}

Distillation is an effective knowledge-transfer technique that uses predicted distributions of a powerful teacher model as soft targets to train a less-parameterized student model. A pre-trained high capacity teacher, however, is not always available. Recently proposed online variants use the aggregated intermediate predictions of multiple student models as targets to train each student model. Although group-derived targets give a good recipe for teacher-free distillation, group members are homogenized quickly with simple aggregation functions, leading to early saturated solutions. In this work, we propose Online Knowledge Distillation with Diverse peers (OKDDip), which performs two-level distillation during training with multiple auxiliary peers and one group leader. In the first-level distillation, each auxiliary peer holds an individual set of aggregation weights generated with an attention-based mechanism to derive its own targets from predictions of other auxiliary peers. Learning from distinct target distributions helps to boost peer diversity for effectiveness of group-based distillation. The second-level distillation is performed to transfer the knowledge in the ensemble of auxiliary peers further to the group leader, i.e., the model used for inference. Experimental results show that the proposed framework consistently gives better performance than state-of-the-art approaches without sacrificing training or inference complexity, demonstrating the effectiveness of the proposed two-level distillation framework.

\end{abstract}

\section{Introduction}

Modeling with tens of millions of parameters helps deep neural networks achieve great success in various applications. However, over-parameterized models are computationally demanding, making them unsuitable for deployment with limited resources or a stringent requirement on latency \cite{denil2013predicting,Han2016DeepMC}. The distillation technique transfers the knowledge of a teacher model in the form of soft predictions to improve the generalization ability of a less-parameterized student model through regularization \cite{ba2014deep,Romero2015FitNetsHF,Yim2017AGF}. Compared to hard ground-truth labels, the soft predicted distributions carry richer information that helps to optimize small networks more effectively \cite{Hinton2015Distil,chaudhari2016entropy}.

The vanilla Knowledge Distillation (KD) method is a two-stage process in which a high capacity teacher model is trained and then
used for distillation. This will increase both training cost and pipeline complexity. Recent attempts on group-based online knowledge distillation explored less costly and unified models to eliminate the necessity of pre-training a large teacher model. \cite{zhang2018deep,Anil2018LargeSD,lan2018knowledge,song2018collaborative}. The main idea is to train a number of student models simultaneously by learning from ground-truth labels as well as distilling from their group-derived soft targets, which is a specific form of aggregation of intermediate peer predictions.

With the absence of a powerful teacher model, the group-derived targets play a key role in transferring group knowledge to each student model. Averaging over predictions of group members is a simple aggregation to derive the targets for representing the group knowledge \cite{zhang2018deep,Anil2018LargeSD,song2018collaborative}. Since the quality of predictions varies among peers, it is important to treat peers unequally \cite{lan2018knowledge}. Unfortunately, naive aggregation functions tend to cause peers to quickly homogenize, hurting the effectiveness of group distillation \cite{kuncheva2003measures,zhou2012ensemble}.
% and changes across iterations
In this paper, we propose a new two-level distillation approach called \textbf{O}nline \textbf{K}nowledge \textbf{D}istillation with \textbf{Di}verse \textbf{p}eers (OKDDip), which involves two types of student models, i.e., multiple auxiliary peers and one group leader. Base distillation is performed among auxiliary peers equipped with a diversity holding mechanism. An ensemble of predictions of these diverse peers is further distilled into the group leader. Unlike naive group-based learning where all peers end up with similar behaviors, trained peer models in our approach could be quite different from each other. It is thus unreasonable to arbitrarily select a single peer for inference. The second-level distillation therefore is necessary to reduce inference cost using only one student model. With this design, OKDDip can be effectively trained without a high capacity teacher network by taking advantages of group distillation and remains efficient for inference.

A key design of OKDDip is that each auxiliary peer assigns individual weights to all the peers during aggregation to derive their own target distributions. We incorporate an attention-based mechanism \cite{vaswani2017attention} to generate a distinct set of weights for each peer to measure the importance of group members. This allows large variation in derived target distributions and hence boosts peer diversity. Note that asymmetric weights exist in our model, which differs from simple aggregation and allows quality peers to excel.

%two peers to weight each other unequally. With this property, it becomes possible to reduce negative effects in one direction and at the same time to accept positive guidance in the other direction, which is critical for group distillation among models with unbalanced performance.

We performed experimental study to evaluate the classification performance of the proposed approach on CIFAR-10, CIFAR-100 and  ImageNet-2012 datasets with a variety of settings based on popular network architectures. Experimental results show that without increasing cost and complexity, our two-level distillation approach consistently generalized better than state-of-the-art online knowledge distillation approaches as well as the classic teacher-guided KD approach. Larger peer diversity and stronger ensemble are observed in our approach when compared to others, demonstrating that the proposed attention-based mechanism works well for diversity holding.

\section{Related Work}

\textbf{Knowledge Distillation.}
KD provides a succinct but effective solution for compressing a pre-trained large teacher model into a smaller student model by steering the student predictions towards teacher predictions \cite{bucilua2006model,ba2014deep,Hinton2015Distil,Polino2018MC}. Compared to hard ground-truth labels, fine-grained class information in soft predictions helps the small model to reach flatter local minima, which results in more robust performance and improves generalization ability \cite{pereyra2017regularizing,keskar2016large}. Several recent works attempt to further improve the performance with new formulations of teacher-learned knowledge \cite{Yim2017AGF,Chen2018Coupled,Ahn2019Variational}.

\textbf{Online Knowledge Distillation.}
Instead of two-stage knowledge transfer, recent work focus on more economic online knowledge distillation without a pre-trained teacher model. Simultaneously training a group of student models by learning from peers' predictions is an effective substitute for teacher-absent knowledge distillation. Some approaches use individual networks with each one corresponding to a student model \cite{zhang2018deep,Anil2018LargeSD}, while some others ask all student models to share the same early blocks to further reduce the training cost \cite{song2018collaborative,lan2018knowledge}. The main difference in these approaches is the way that each student model learns from others. In \cite{zhang2018deep}, each student model learns from the simple average of predictions of all other group members and requires complex asynchronous updating among different networks. A similar variants called co-distillation investigated the potential benefits for online distillation in the distributed learning \cite{Anil2018LargeSD}.  In \cite{lan2018knowledge}, all student models share the same target distribution by averaging predictions of all the members with weights learned by a fully connected layer. Unfortunately, simply treating each peer to be equally important or forcing all the members to learn from the same targets would hurt the diversity among students, which limits the effectiveness of within-group knowledge transfer.

\textbf{Self-Attention.}
Attention was introduced in natural language processing for encoding each word with others which are most relevant regarding to the target task \cite{BahdanauCB2015Neural}. It has been also successfully applied to other types of data such as image and graph with state-of-the-art performance. In particular, self-attention or intra-attention refers to the mechanism of capturing global dependencies by calculating the response at a position through attending to all the neighbors \cite{vaswani2017attention}. Specifically, the input representation of a position, such as a word in a sequence \cite{vaswani2017attention}, a pixel within an image region \cite{zhang2018self} or a node in a graph \cite{velivckovic2017graph}, is linearly mapped into three vectors called \textit{query}, \textit{key}, and \textit{value}. The output of this position is obtained by averaging the \textit{values} of its neighbors, e.g., words in the same sequence, with different weights, which are calculated by matching the \textit{query} of this position to the \textit{keys} of neighbors.

\section{Online Knowledge Distillation with Diverse Peers}

\subsection{Learning with labels}
Suppose we are dealing with classification tasks with a labeled dataset $\mathcal{D}=\{(x_i,y_i)\}^n_i$. The goal of learning is to find a parameterized mapping $f(x,\boldsymbol{\theta}): \mathcal{X} \mapsto [0,1]^{|\mathcal{Y}|}$, which can generalize to unseen data.
A classifier is typically trained by minimizing the cross entropy between the predicted class probabilities $\boldsymbol{q}_i$ and the one-hot ground-truth label distribution $\boldsymbol{l}_i$ of each training sample
\begin{equation}
\mathcal{L}_{gt}=-\sum_{i,j} l_{ij}\log q_{ij},
\end{equation}
where $l_{ij}=1$ if $y_i=j$, and 0 otherwise. $\boldsymbol{q}_i=\sigma(\boldsymbol{g}_i,T)$ is calculated with softmax of logits $\boldsymbol{g}_i$, i.e., outputs of the last fully connected layer, i.e.,
\begin{equation}
\label{eq:qic}
q_{ij}=\frac{\exp(g_{ij}/T)}{\sum_k \exp(g_{ik}/T)},
\end{equation}
and parameter $T$ is usually set to 1.

Next, we present the detailed formulation of the proposed approach after a brief review of teacher-required knowledge distillation.
\subsection{Distillation with a teacher model}
In teacher-presented knowledge distillation, a student model is trained by the teacher-predicted soft distribution $\boldsymbol{t}$ together with hard ground-truth labels. The soft distribution or targets $\boldsymbol{t}$ are calculated with softmax of logits as Equation (\ref{eq:qic}) with a temperature $T>1$. A higher $T$ means softer distributions. Following the suggestion in \cite{lan2018knowledge}, we set $T$ to 3 in this paper for all methods.

Knowledge is transfered by aligning the student predicted distribution $\boldsymbol{q}'$ also after softmax with the same temperature (i.e., $T=3$) to the target distribution. Specifically, Kullback-Leibler (KL) divergence between  $\boldsymbol{t}$ and $\boldsymbol{q}'$ may be used to define the distillation loss
\begin{equation}
\mathcal{L}_{dis}= KL(\boldsymbol{t},\boldsymbol{q}')=\sum_{i,j} t_{ij} \log \frac{t_{ij} }{q'_{ij}}.
\end{equation}
With both hard and soft labels, the total loss for training a teacher-presented distillation model is
\begin{equation}
\mathcal{L}_{KD}= \mathcal{L}_{gt}+T^2\mathcal{L}_{dis},
\end{equation}
where $\mathcal{L}_{dis}$ is multiplied by $T^2$ before combination to ensure that the contribution of distillation term keeps roughly unchanged if the temperature is changed \cite{Hinton2015Distil}. It is worthy to note that the predicted probabilities of the student model $\boldsymbol{q}$ are computed from the logits with temperature $T=1$ when aligning with hard ground-truth labels but with a higher temperature when aligning with soft targets in distillation training. To be more clear, we use $\boldsymbol{q}$ for the $T=1$ version, and $\boldsymbol{q}'$ for the high temperature version through out this paper.

\begin{figure*}
	\centering
	\includegraphics[width=0.9\textwidth]{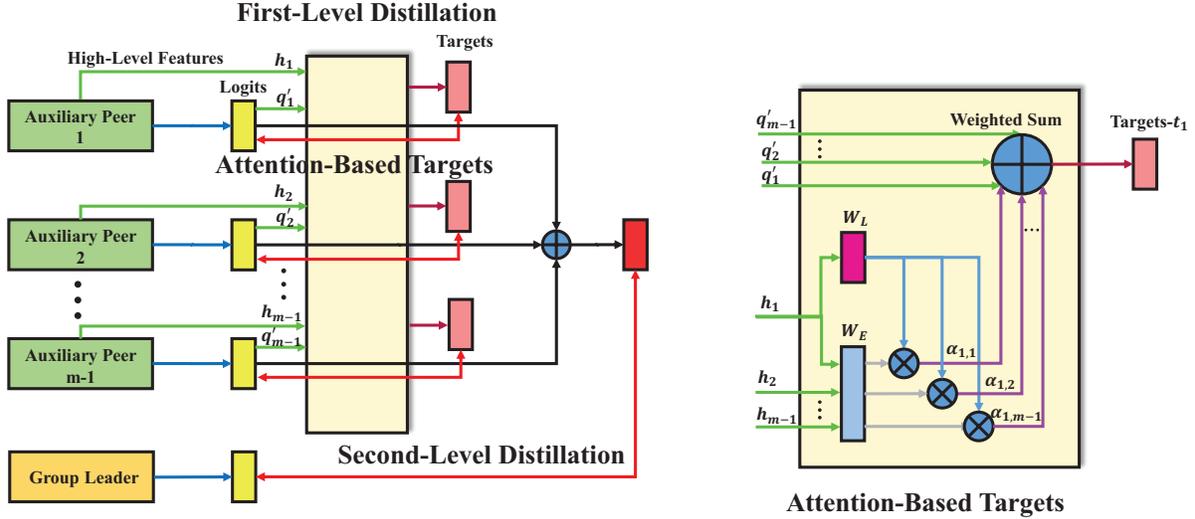}
	\caption{An overview of the proposed Online Knowledge Distillation with Diverse Peers (OKDDip). (Left) Two-level distillation. The first-level distillation performs group-based learning among $m-1$ auxiliary peers by learning from their own targets. The second-level distillation transfers group knowledge of diverse peers to the group leader, i.e., the final model for deployment. (Right) Attention-based targets derivation. The targets of an auxiliary peer is computed as weighted sum of logits of group members. Each weight $\alpha_{ab}$ for auxiliary peer $a$ to attend member $b$ in deriving its targets is calculated with normalized Embedded Gaussian distance between their mapped features $\mathbf{W}_L^T\boldsymbol{h}_a$ and $\mathbf{W}_E^T\boldsymbol{h}_b$, where  $\mathbf{W}_L$ and $\mathbf{W}_E$ are two linear projection matrices, $\boldsymbol{h}_a$ and $\boldsymbol{h}_b$ are original high-level features. }\label{fg:model}
\end{figure*}

\subsection{Two-level distillation}
\label{sec:two}
The proposed two-level distillation framework for group-based knowledge distillation is illustrated in Figure \ref{fg:model}. All the $m$ student models including $m-1$ auxiliary peers and one group leader use the same network architecture, which consists of a feature extractor to produce high-level features followed by a classifier to produce logits. 
%In a multi-branch setting, the $m$ student models share the first several blocks of layers. 
For convenience of denotation, all the student models are indexed from 1 to $m$, where 1 to $m-1$ denote the auxiliary peers and $m$ corresponds to the group leader. The predicted distribution (i.e., the high temperature version) of the $a$th student model is denoted as $\boldsymbol{q}'_a$ with $a=1,\ldots m$.

\textbf{Loss function.} For the first-level distillation, each auxiliary peer for $a=1,2,\ldots m-1$ distills from its own group-derived soft targets $\boldsymbol{t}_a$, which is computed by aggregating predictions of all peers with different weights
\begin{align}
\boldsymbol{t}_{a}=\sum_{b=1}^{m-1}\alpha_{ab} \cdot \boldsymbol{q}'_{b},
\end{align}
where $\alpha_{ab}$ represents the extent to which the $b$th member is attended in derivation $\boldsymbol{t}_{a}$, and $\sum_{b}\alpha_{ab}=1$. We will elaborate on these attention-based weights later.
The distillation loss of all auxiliary peers is then given as
\begin{equation}
\mathcal{L}_{dis1}=  \sum^{m-1}_{a=1} KL(\boldsymbol{t}_a,\boldsymbol{q}'_a), \label{eq:dis1}
\end{equation}
which could be regarded as weighted regularization for the output distribution. As pointed in \cite{pereyra2017regularizing}, penalizing the confident prediction could prevent over-fitting by increasing the probabilities assigned to incorrect classes.

The group knowledge of those auxiliary peers is distilled further to the group leader (i.e., the $m$th student model) with the second distillation
\begin{equation}
\mathcal{L}_{dis2}=   KL(\boldsymbol{t}_{m},\boldsymbol{q}'_{m}),
\end{equation}
which is similar to the classic KD process in the way of transferring the knowledge of an ensemble to a student model, but in online fashion. With diversity enhancement by the first-level distillation, we simply average the predictions of all auxiliary peers to compute $\boldsymbol{t}_{m}$.

The overall loss of the proposed approach is given as
\begin{equation}
\mathcal{L}_{OKDDip}= \sum^m_{a=1}\mathcal{L}_{gt}(a) +  T^2\mathcal{L}_{dis1} + T^2\mathcal{L}_{dis2}, \label{eq:L_OKDDip}
\end{equation}
where the first term is the total cross entropy loss to the ground-truth labels of all the $m$ student models\footnote{Following \cite{lan2018knowledge,laine2016temporal}, the two distillation terms would be multiplied by an iteration-dependent weighting function during implementation to avoid large contribution of distillation in early stages.}.

\textbf{Attention-based weights.}
With different initializations, quality of intermediate predictions varies among peer models, which should contribute different extents in deriving the soft target distributions for each of the auxiliary peers. Simply treating all peers equally would make the distilled model suffer from negative contributions from low quality predictions. We expect the weights to capture the relative importance of peers to a distilled model.

Inspired by the self-attention mechanism \cite{vaswani2017attention}, we project the extracted features of each peer model $\boldsymbol{h}_{a}$ into two subspaces separately by linear transformation
\begin{align}
L(\boldsymbol{h}_{a})=\mathbf{W}_{L}^{T}\boldsymbol{h}_{a} \quad and \quad
E(\boldsymbol{h}_{a})=\mathbf{W}_{E}^{T}\boldsymbol{h}_{a}, \label{eq:att}
\end{align}
where $\mathbf{W}_{L}$ and $\mathbf{W}_{E}$ are the learned projection matrices shared by all auxiliary peers.
%Linear transformation is used as the extracted feature space is expected to be linearly separable due to the fact that a following linear classifier usually suffices in practice. 
Similar to self-attention, $\alpha_{ab}$ is calculated as Embedded Gaussian distance with normalization
\begin{align}
\alpha_{ab}=\frac{e^{L(\boldsymbol{h}_{a})^{T}E(\boldsymbol{h}_{b})}}{\sum_{f=1}^{m-1} e^{L(\boldsymbol{h}_{a})^{T}E(\boldsymbol{h}_{f})}}.
\end{align}

The two separate transformation matrices may capture different semantic information. Weights generated in this way have the following merits:
\begin{itemize}
	\item \textbf{Asymmetric:} The asymmetric property provides a possible way to suppress negative effect in one direction without stopping positive guidance in the other, which is important for mutual learning between two peers optimized to different levels. It on the one hand reduces the extent to which a well-behaved model is affected by a poorly performed peer by assigning a small weight, and on the other hand allows the less optimized model to learn from the better optimized one with a large weight.
	\item \textbf{Dynamic:} The performance of peer models changes during training, updating weights in iterations allows each model to attend a dynamic set of peers adaptively.
\end{itemize}
It is shown in our experimental study that the above properties enable OKDDip to outperform state-of-the-art approaches as well as its simplified variants.

\textbf{Why learning with distinct target distributions leads to diversity?}
In the first-level distillation, each auxiliary peer has an individual set of weights to measure the extent to which it attends all group members from its own point of view. Such kind of a personalized aggregation increases the independency between soft target distributions of different peers, which is helpful for alleviating diversity degradation during group-based distillation as demonstrated in experiments later. Next, we provide analytical discussions on why $\mathcal{L}_{dis1}$ leads to peer diversity based on its approximation given below.

If the logits are zero-meaned before computing $\boldsymbol{q}'$, the KL-divergence distillation loss in Equation (\ref{eq:dis1}) then is approximated by the Mean-Square-Error \cite{Hinton2015Distil}:
\begin{equation}
\label{eq:dis1_new}
\mathcal{L}'_{dis1}=  \frac{1}{2} \sum_{a=1}^{m-1}  \|\boldsymbol{q}'_a-\boldsymbol{t}_a  \|^2 = \frac{1}{2} \sum_{a=1}^{m-1} \|\boldsymbol{q}'_{a}-\sum_{b=1}^{m-1}\alpha_{ab}\boldsymbol{q}'_{b}  \|^2.
\end{equation}
The above loss function is minimized with $\alpha_{aa}=1$ and $\alpha_{ab}=0$ for all $b\neq a$, which means that each peer does not learn from others at this point. This could be regarded as an extreme of our approach where all auxiliary peers are trained independently without any group-based distillation. Incorporated with such kind of loss into the total objective function in Equation (\ref{eq:L_OKDDip}), the proposed approach is able to keep a proper balance between group sharing and independent learning, which allows it to leverage the information distilled from other peers while preventing quick diversity diminishing.

\textbf{Training and deployment.}
The proposed framework may be implemented with branch-based or network-based student models. In a branch-based setting, all student models share the first several layers to use the same low-level features, and separate from each other from a certain layer to have individual branches for high-level feature extraction and classification. In a network-based setting, student models are individual networks.
%The training process uses standard optimization procedure with stochastic gradient descent to train all the student models simultaneously to achieve faster convergence and better performance \cite{mescheder2017GANs,Vaishnavh2017Gradient}. 
All auxiliary peers are discarded after training and only the group leader is kept for deployment. There is no additional increase in complexity or cost compared to other group-based approaches given the number and architecture of student models are the same.

\begin{table*}[htbp]
	\centering
	\caption{Error rates (Top-1, \%) on CIFAR-10. OKDDip: network-based (1st column) and branch-based (2nd column).}
	\label{Table:CIFAR-10}
	\begin{tabular}{cccccc|cc}
	\toprule
 	Network & Baseline & Ind & DML & CL-ILR & ONE & \multicolumn{2}{c}{OKDDip} \\
	\midrule
	DenseNet-40-12
	& 6.87 $\pm$ 0.02 & 6.97 $\pm$ 0.03 & 6.50 $\pm$ 0.02 & 7.02 $\pm$ 0.08 & 6.85 $\pm$ 0.15 & \textbf{5.94 $\pm$ 0.05} & 6.48 $\pm$ 0.12 \\
	\midrule 
	ResNet-32
	& 6.34 $\pm$ 0.03 & 5.99 $\pm$ 0.15 & 6.18 $\pm$ 0.05 & 6.06 $\pm$ 0.07 & 5.94 $\pm$ 0.06 & 5.62 $\pm$ 0.07 & \textbf{5.58 $\pm$ 0.08}\\
	\midrule
	VGG-16
	& 6.12 $\pm$ 0.15 & 6.03 $\pm$ 0.01 & 5.94 $\pm$ 0.04 & 6.22 $\pm$ 0.10 & 6.16 $\pm$ 0.08 & \textbf{5.88 $\pm$ 0.04} & \textbf{5.87 $\pm$ 0.03} \\
	\midrule
	ResNet-110
	& 5.46 $\pm$ 0.02 & 4.95 $\pm$ 0.02 & 5.68 $\pm$ 0.03 & 4.88 $\pm$ 0.12 & 5.02 $\pm$ 0.04 & \textbf{4.54 $\pm$ 0.07} & \textbf{4.56 $\pm$ 0.11} \\
	\midrule
	WRN-20-8
	& 5.27 $\pm$ 0.06 & 5.35 $\pm$ 0.02 & 5.04 $\pm$ 0.08 & 5.12 $\pm$ 0.16 & 5.29 $\pm$ 0.02 & \textbf{4.84 $\pm$ 0.07} &  5.06 $\pm$ 0.04 \\
	\bottomrule
	\end{tabular}		
\end{table*}

\section{Experiments}

We provide experimental results in this section to evaluate the performance of the proposed approach for image classification. In addition to the overall generalization ability, we also study the diversity maintenance ability for group-based distillation and conduct several ablation studies on the attention mechanism as well as the two-level strategy. Finally, we analyze the impact of group size, i.e., the number of student models, and extend our method with an additional pre-trained teacher model. All evaluations are made in comparison with state-of-the-art approaches. More results are provided in supplementary materials. 

% We use PyTorch-1.0 for implementation and conduct all experiments on an NVIDIA TITAN Xp GPU. 

\textbf{Datasets and Architectures.} Three image classification datasets are used in the following evaluations. CIFAR-10 and CIFAR-100 \cite{krizhevsky2009learning} both contain 50,000$\slash$10,000 training$\slash$testing colored natural images with $32\times 32$ pixels, which are drawn from 10$\slash$100 classes. ImageNet-2012 \cite{Russakovsky2015ImageNet} is a more challenging dataset consisting of about 1.3 million training images and 50 thousand validation images from 1000 classes. We adopted a standard augmentation procedure as \cite{he2016deep,huang2017densely,zhang2018deep}. For preprocessing, we normalized all images by channel means and standard deviations. The size of each training sample is  32$\times$32 for CIFAR-10/100 and 224$\times$224 for ImageNet-2012. Six network architectures are used in our experiments, namely VGG-16 \cite{simonyan2014very}, ResNet-32, ResNet-34, ResNet-110 \cite{he2016deep}, WRN-20-8 \cite{zagoruyko2016wide}, and DenseNet-40-12 \cite{huang2017densely}. 
 
\textbf{Settings.} We use stochastic gradient descent with Nesterov momentum for optimization and set the initial learning rate to 0.1, momentum to 0.9. For CIFAR-10$\slash$CIFAR-100 dataset, we set the mini-batch size to 128 and weight decay to $5\times 10^{-4}$. The learning rate is divided by 10 at 150 and 225 of the total 300 training epochs for these two datasets. For ImageNet-2012 dataset, we set the mini-batch size to 256, the weight decay to $1\times 10^{-4}$, and the learning rate is divided by 10 at 30 and 60 of the total 90 training epochs. All results are reported in means (standard deviations) over 3 runs. Codes are available at \url{https://github.com/DefangChen/OKDDip-AAAI2020}.
 
\textbf{Approaches compared.} We compare the proposed OKDDip to several recently proposed online knowledge distillation approaches, including network-based DML \cite{zhang2018deep}, branch-based CL-ILR \cite{song2018collaborative}, and ONE \cite{lan2018knowledge}. The ``Baseline'' approach trains a model by ground-truth labels only and ``Ind'' refers to the degenerated branch-based approach that trains each student model individually without any group distillation. The classic teacher-guided KD approach \cite{Hinton2015Distil} and a high capacity teacher model are also included for comparison. 

For branch-based student models, all student models share the first several blocks of layers and separate from the last block for CIFAR-10/100 and the last two blocks for ImageNet-2012 to form a multi-branch structure as \cite{lan2018knowledge}. All group-based online knowledge distillation approaches are compared with student models of the same architecture and the same number. Results of group-based approaches are generated with 4 student models except those which are regarding the impact of group size. Since we keep one student model to serve as group leader in OKDDip, the number of peers in group-based aggregation is one less than that of DML, CL-ILR and ONE.

\begin{table*}[htbp]
	\centering
	\caption{Error rates (Top-1, \%) on CIFAR-100. OKDDip: network-based (1st column) and branch-based (2nd column).}
	\label{Table:CIFAR-100}
	\resizebox{0.95\textwidth}{!}{
	\begin{tabular}{cccccc|cc}
		\toprule
		Network & Baseline & Ind & DML & CL-ILR & ONE & \multicolumn{2}{c}{OKDDip} \\
		\midrule
		DenseNet-40-12
		& 28.97 $\pm$ 0.23 & 29.20 $\pm$ 0.09 & 26.64 $\pm$ 0.17 & 28.61 $\pm$ 0.12 & 28.76 $\pm$ 0.18 & \textbf{26.10 $\pm$ 0.03} & 28.34 $\pm$ 0.02 \\
		\midrule
		ResNet-32
		& 28.76 $\pm$ 0.08 & 27.84 $\pm$ 0.05 & 26.47 $\pm$ 0.26 & 27.44 $\pm$ 0.05 & 26.50 $\pm$ 0.13 & \textbf{25.40 $\pm$ 0.08} & 25.63 $\pm$ 0.14 \\	
		\midrule
		VGG-16
		& 26.19 $\pm$ 0.12 & 25.81 $\pm$ 0.18 & 25.33 $\pm$ 0.03 & 25.62 $\pm$ 0.11 & 25.63 $\pm$ 0.03 & \textbf{24.88 $\pm$ 0.06} & 25.15 $\pm$ 0.19 \\
		\midrule
		ResNet-110
		& 24.12 $\pm$ 0.20 & 23.54 $\pm$ 0.15 & 22.50 $\pm$ 0.11 & 21.56 $\pm$ 0.09 & 21.67 $\pm$ 0.12 & \textbf{21.09 $\pm$ 0.17} & 21.14 $\pm$ 0.14 \\
		\midrule
		WRN-20-8
		& 22.50 $\pm$ 0.44 & 21.85 $\pm$ 0.12 & 20.21 $\pm$ 0.11 & 20.44 $\pm$ 0.13 & 21.19 $\pm$ 0.12 & \textbf{19.63 $\pm$ 0.07} & 20.06 $\pm$ 0.05\\
		\bottomrule
	\end{tabular}
}
\end{table*}

\subsection{Comparison of classification error rates}

Table \ref{Table:CIFAR-10} and Table \ref{Table:CIFAR-100} give the Top-1 classification error rates (\%) on CIFAR-10 and CIFAR-100 based on five different network architectures with parameters range from 0.18M (DenseNet-40-12) to 15.30M (VGG-16). Following the original papers, the results of compared methods are the averaged accuracy of different student models, and the results of OKDDip are generated by the group leader. Two groups of results are reported for OKDDip with network-based (1st column) and branch-based (2nd column) implementation.

From these two tables, it is shown that OKDDip achieved lower error rates than all other approaches in both types of student models. Specifically, for the network-based setting, OKDDip outperformed the ``Baseline'' and DML by 17\% and 20\% in the best case (ResNet-110), showing that the two-level distillation strategy with attention-based mechanism is more effective than existing ones for group learning. It is also seen that OKDDip achieved lower error rates than ``Ind'', CL-ILR and ONE by 16\%, 8\% and 9\% in the best case (ResNet-32/110), respectively, showing that our framework remains to be effective with the multi-branch setting. Generally, OKDDip gives slightly better results with network-based student models, which have more independent parameters to give a larger room to maintain the diversity of peers. We also found that the compared group-based online approaches consistently achieved better performance than the ``Baseline'' and ``Ind'' on CIFAR-100 but sometimes fall behind on CIFAR-10, especially for VGG-16 network architecture, which indicates that the homogenization problem tends to become even more severe for dealing with easier dataset.   

%We also test the performance of each method for large-scale image classification datasets. As show in Table \ref{Table:ImageNet}, either branch-based or network-based student models, OKDDip both outperform the compared approaches by XX\%.

Table \ref{Table:ImageNet} gives the classification error rates for the Image-\\Net-2012. We obtain similar observations as above. For the following experiments, results of OKDDip are generated based on the branch-based setting.

\begin{table}
	\caption{Error rates (Top-1, \%) for ResNet-34 on ImageNet-2012. OKDDip: network-based (1st column) and branch-based (2nd column).}
	\label{Table:ImageNet}
	\centering
	\begin{tabular}{cccc|cc}
		\toprule
		Baseline & DML & CL-ILR & ONE & \multicolumn{2}{c}{OKDDip} \\
		\midrule
		26.76 & 26.03 & 26.06 & 25.92 &  \textbf{25.42} & 25.60 \\
		\bottomrule
	\end{tabular}
\end{table}

\subsection{Diversity analysis}

Next, we evaluate whether OKDDip produces more diverse student models compared with other group-based approaches. For each method, the diversity is measured by the average Euclidean distance between the predictions of each pair of peers. 

Figure \ref{fig:diversity} plots the results of peer diversity comparison with four approaches for CIFAR-100. The diversity of ``Ind'' can be regarded as an upper bound since the student models are trained independently. 

As shown in Figure \ref{fig:diversity}, peer diversity for all the approaches are very small when starting from random initialization and climb rapidly after several optimization steps. However, it drops quickly for CL-ILR and ONE until the learning rate is decreased (at the 150 epoch), from which it climb again but with a slower speed. 
%the diversity of peers trained by each method both are quite small at first for the randomly initialized parameters and climb up rapidly after several optimization steps. Then, the diversity will generally drop during same learning rate schedule and increase once learning rate decrease (At 150 and 225 epochs). %which is similar to the error change during training.
Through the whole epochs, the diversity of peers trained by OKDDip is significantly larger than CL-ILR and ONE for both the two ResNet architectures, approaching to those of ``Ind'', especially for ResNet-110. This demonstrates that the self-attention mechanism in proposed two-level distillation framework works successfully for alleviating homogenization during group-based distillation. 

\begin{figure}
	\centering
	\subcaptionbox{ResNet-32\label{fig:diversity_32}}{
		\includegraphics[width=0.48\linewidth]{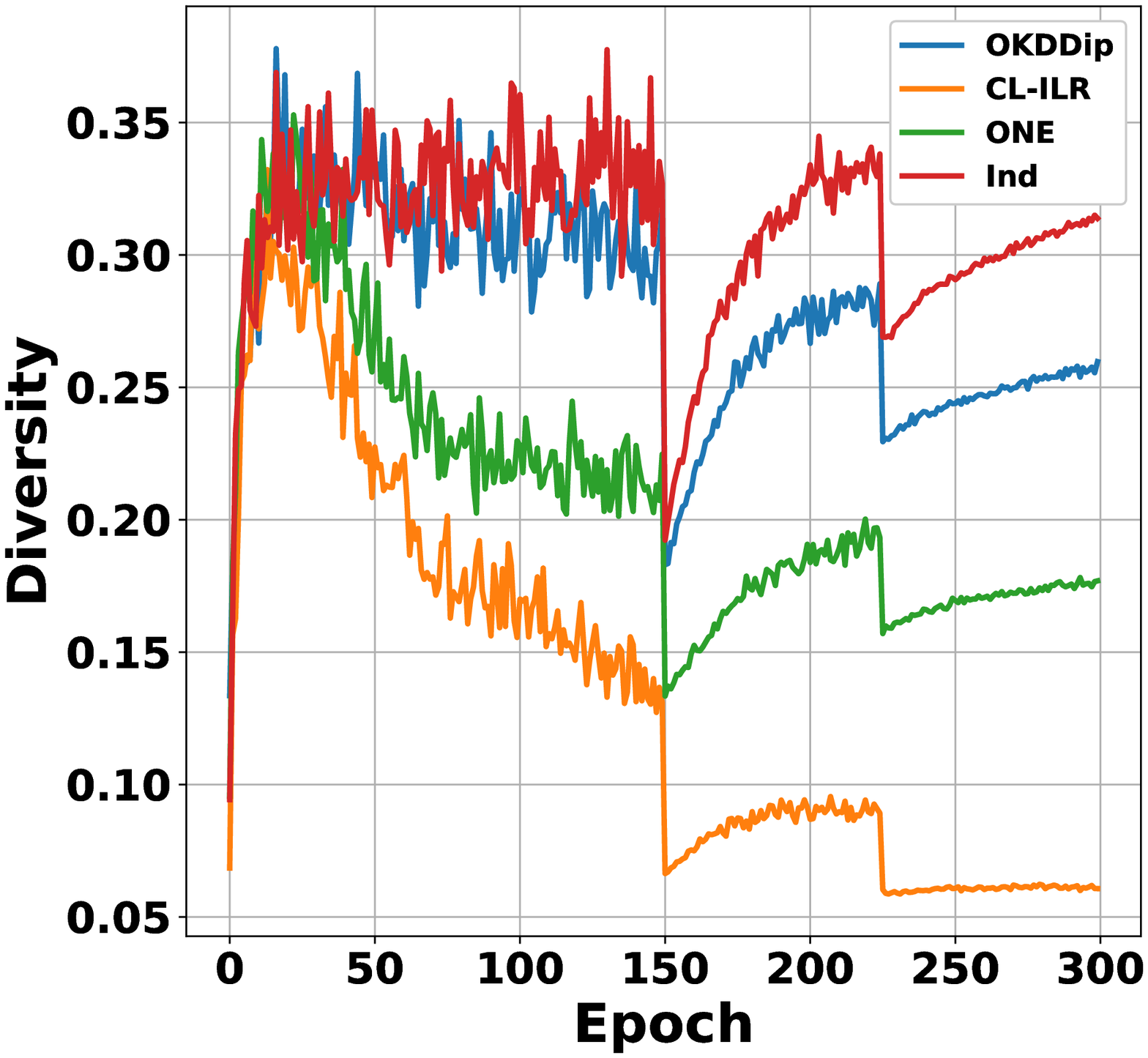}}
	%\hspace{0.1\textwidth}
	\subcaptionbox{ResNet-110\label{fig:diversity_110}}{
		\includegraphics[width=0.48\linewidth]{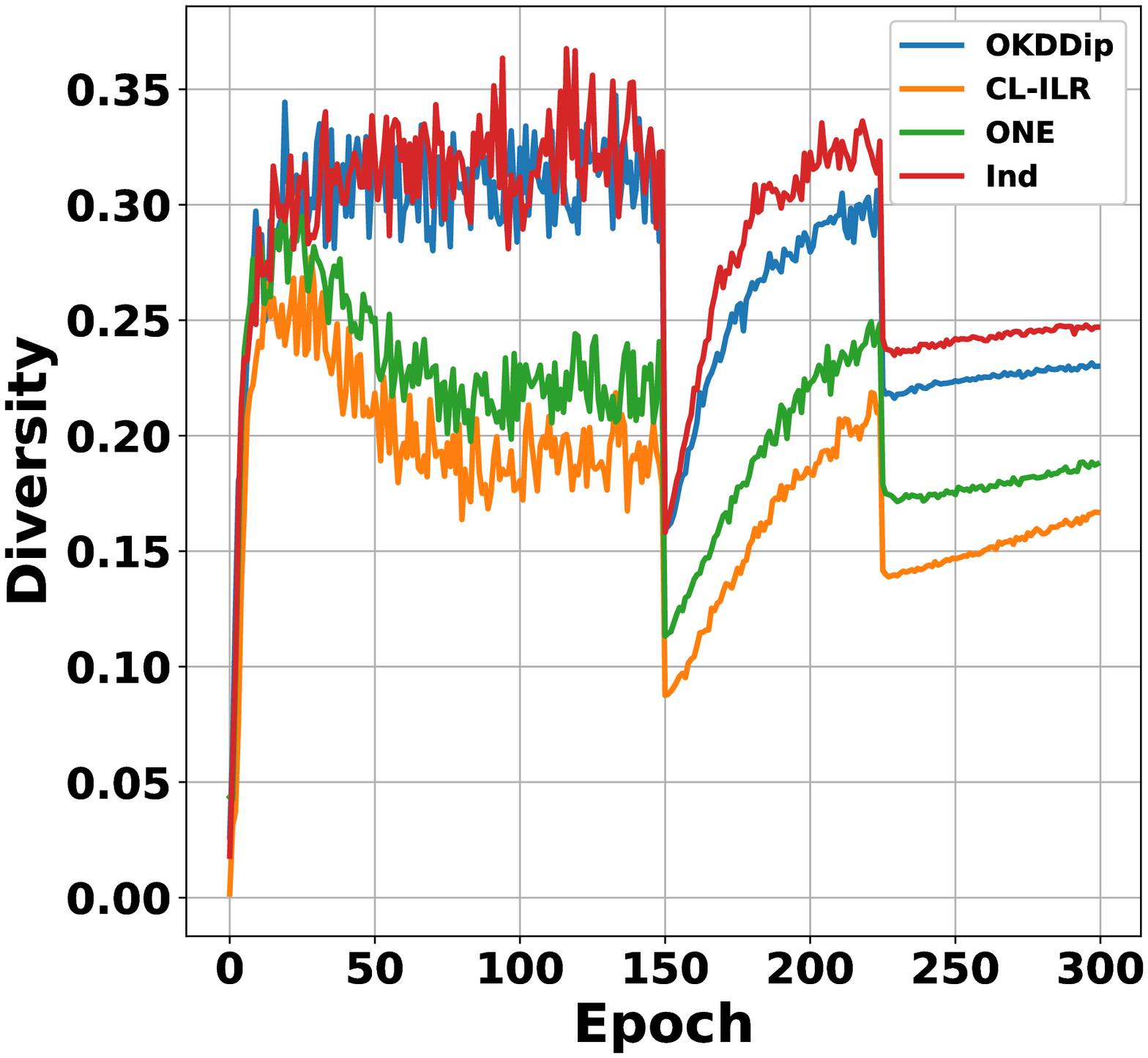}}
	\caption{Peer diversity comparison with branch-based models during training on CIFAR-100.}
	\label{fig:diversity}
\end{figure}

To show whether diverse peers lead to a stronger ensemble, we also evaluate the accuracy of averaged predictions of trained peers, which can represent how powerful the generated group knowledge to some extent. Therefore, only auxiliary peers are counted in for OKDDip. 
%Table \ref{BranchE-Table} gives the ensemble accuracy for OKDDip (3 auxiliary peers) and compared methods (4 peers).

From Table \ref{BranchE-Table}, thanks to large peer diversity, more effective ensemble is observed with OKDDip,
%with one less peers which is kept as the group leader, 
which even achieve lower error rates than the ensemble of individually trained student models. Together with Table \ref{Table:CIFAR-100} and Figure \ref{fig:diversity}, we can find that the compared method CL-ILR and ONE improve the accuracy of individual student models ranging from 0.7\% to 8\% while decreasing the diversity among different peers, which sacrifices the effectiveness of ensemble models. The error rates even increase by 14\% and 5\% for ResNet-32. But OKDDip can learn more distinct targets for each auxiliary peer and thus maintain the diversity, which further benefit the learning for group leader, leading to the success for both individual and ensemble models. It is also shown that diversity only is not enough to ensure a good ensemble when comparing with the results of ``Ind''.

%Effective group distillation such as the proposed two-level framework is important to enhance the performance of each peer.

\begin{table}
	\caption{Error rates (Top-1, \%) of ensemble predictions with branch-based student models on CIFAR-100.}
	\label{BranchE-Table}
	\centering
	\begin{tabular}{cccc|c}
		\toprule
		Network & CL-ILR & ONE & OKDDip & Ind \\
		\midrule
		VGG-16      & 25.56 & 25.54 & \textbf{24.95} & 25.62 \\
		ResNet-32 	& 27.01 & 24.90 & \textbf{23.45} & 23.74 \\
		ResNet-110  & 20.19	& 20.14	& \textbf{19.54} & 20.18 \\
		\bottomrule
	\end{tabular}
\end{table}

\subsection{Ablation Study}

\begin{table*}
	\caption{Ablation study: Error rates (Top-1, \%) for ResNet-32 on CIFAR-100}
	\label{Table-A}
	\centering
	\begin{tabular}{cccccc}
		\toprule
		w/o SA (random) & w/o SA (entropy) & w/o SA (mean) & w/o SA (asymmetry) & w/o two-level & OKDDip \\
		\midrule
		28.24 $\pm$ 0.16 & 26.71 $\pm$ 0.19 & 26.35 $\pm$ 0.14 & 26.05 $\pm$ 0.17 & 27.79 $\pm$ 0.14 & \textbf{25.63 $\pm$ 0.14} \\
		\bottomrule
	\end{tabular}
\end{table*}		
\begin{table*}[htbp]
	\caption{Error rates (Top-1, \%) for ResNet-32 with an additional teacher}
	\label{KD-Table}
	\centering
	\begin{tabular}{ccccc|c}
		\toprule
		DataSet  & Baseline & KD & OKDDip & OKDDip+KD & Teacher \\
		\midrule
		CIFAR-10 
		& 6.34 $\pm$ 0.03 & 6.08 $\pm$ 0.11 & 5.58 $\pm$ 0.08 & \textbf{5.36 $\pm$ 0.06} & 5.27 $\pm$ 0.23 \\
		CIFAR-100 
		& 28.76	$\pm$ 0.08 & 26.51 $\pm$ 0.14 & 25.63 $\pm$ 0.14 & \textbf{24.92 $\pm$ 0.08} & 24.12 $\pm$ 0.20 \\
		\bottomrule
	\end{tabular}
\end{table*}
To further show the benefit of each individual OKDDip components, especially for the self-attention (SA) mechanism, we perform various ablation studies on CIFAR-100 based on ResNet-32. Specifically, we compare the performance of OKDDip with the following five ways of ablations. 

(1) w/o SA (random). A random attention matrix with normalization is used to put randomly generated belief among peers. This leads to higher error rates by 2.61\% (28.24\%-25.63\%).

(2) w/o SA (entropy). In order to validate the effectiveness of dynamic weight modeling of SA, we completely remove $\mathcal{L}_{dis1}$ from objective function and let the peers only learn from $\mathcal{L}_{gt}$ with an entropy term, i.e., each peer attends only to itself, which reduces the performance by 1.08\%.

(3) w/o SA (mean). Simple average is applied to aggregate the predictions of peers in the first-level distillation. This causes 0.72\% performance drop due to quick homogenization. 

%为什么two-level让它比其他方法好？
%Simply using average predictions of peers to derive soft targets for the first-level distillation will cause 0.72\% accuracy drop due to quick homogeneity, but this result still outperforms the compared methods for the two-level framework.

(4) w/o SA (asymmetry). Another special case that ablates asymmetry of SA by forcing $\mathbf{W}_{E}$ and $\mathbf{W}_{L}$ as identity matrices. The weaker performance (0.42\% error rate increase) indicates that the asymmetry merit of OKDDip indeed helps to alleviate more the negative effect from the poor-optimized models to the well-behaved models during training.		

%Moreover, OKDDip performed consistently better than OKDDip-S, indicating that the attention-based mechanism which produces asymmetric weights, helps to obtain further improvements.

(5) w/o two-level. The second-level distillation is ablated by removing the group leader from training and inference with a randomly chosen student model. This increases the error rate by 2.16\%, which confirms the usefulness of the second-level distillation.

%Ablating the two-level distillation process by dropping the group leader from training and randomly pick one student model out results in the performance decreased by 2.16\%, which confirm the necessity of distinguishing the role of student models. 

\subsection{Impact of the group size}
\begin{figure}
	\centering
	\includegraphics[width=0.45\textwidth]{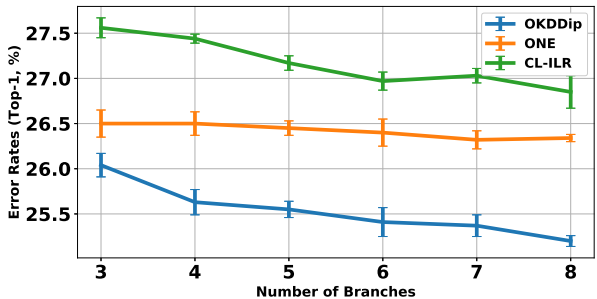}
	\caption{Impact of group size with branch-based ResNet-32 on CIFAR-100.}
	\label{fig:Branch}
\end{figure}
In this section, we evaluate the impact of the group size on the performance of group-based distillation approaches. We compare OKDDip with ONE and CL-ILR using ResNet-32 with the branch-based setting. Figure \ref{fig:Branch} plots the error rates of the three approaches on CIFAR-100 with respect to the total number of branches changing from 3 to 8.

It is seen that OKDDip performs the best among the three in all the cases. OKDDip has a curve that sloped down more sharply, which means that there is still relative large room for further improvement if a larger group size is allowed in deployment. Although ONE gave better results than CL-ILR, further improvement by adding more branches
indicates that increasing the group size does not help much due to the homogenization problem.

\subsection{When a teacher is available}

Although our approach is mainly designed for teacher-free deployment, it is still interesting to know that whether it can be further improved if a pre-trained teacher is indeed available. Here, we use ResNet-110 as the teacher model, the multi-branch ResNet-32 as student models of OKDDip. To make a teacher-guided OKDDip, denoted as OKDDip+KD, the KL-divergence losses between the predictions of each student and the teacher are added into the original loss function of OKDDip for optimization.

Table \ref{KD-Table} gives the results of OKDDip+KD compared with the ``Baseline'', the classic KD approach, and the teacher-free OKDDip. We also included the results of the teacher model.
It is observed that among all the four approaches with a small student model for inference, OKDDip+KD becomes the most competitive one, which approaches the level of the teacher model and outperforms ``Baseline'' by 15\% and 13\% on CIFAR-10 and CIFAR-100 respectively. This shows that a powerful teacher is still helpful to further improve the generalization ability of OKDDip.
The effectiveness of using a teacher is also demonstrated by comparing KD and the ``Baseline''. More importantly, it is observed that with group-based two-level online distillation, the teacher-absent OKDDip already outperformed the teacher-assisted KD with significant improvement.

\section{Conclusion}

Group-based knowledge distillation is a good substitute for knowledge transfer when a pre-trained high capacity model is not easily accessible. It is critical but challenging for group learning without too much diversity diminishing among peers. We proposed a novel two-level framework for effective online distillation. The base distillation works as diversity maintained group distillation with several auxiliary peers, which are discarded after training. The second-level distillation transfers the diversity enhanced group knowledge to the ultimate student model called group leader. Experimental results show that by distilling from distinct target distributions derived with weights from an attention-based component, peer diversity is maintained to a relative large extent throughout group learning, leading to effective online knowledge transfer. This finally makes the proposed approach outperforms the state-of-the-art online knowledge distillation approaches without additional training and inference cost. Our results also show that although a teacher model is still helpful, our teacher-free OKDDip already achieves higher accuracy by a large margin than the teacher-presented KD model, making it a promising competitive choice for deployment.

\bibliography{MC}
\bibliographystyle{aaai}

%\newpage

%\vskip 0.1in
%\hrule height 4pt
%\vskip 0.25in
%\vskip -\parskip
%\begin{centering}
%	\begin{bf}
%		\begin{LARGE}
%			Supplementary Material
%		\end{LARGE}
%	\end{bf}
%\par
%\end{centering}
%\vskip 0.29in
%\vskip -\parskip
%\hrule height 1pt
%\vskip 0.09in%
%\vskip 0.3in

\section{Supplementary Material}
\subsection{Diversity Analysis}
This section provides additional experimental results. We obtain similar observations as the results in the main submission.

\subsection{Ensemble Results}

Table \ref{NetworkE-C10}-\ref{BranchE-C10} give the ensemble results for network-based student models on CIFAR-10, network-based student models on CIFAR-100, branch-based student models on CIFAR-10, respectively. 

Table \ref{KDE-Table} gives the ensemble results of OKDDip+KD compared with the ``Baseline'', the classic KD approach, and the teacher-free OKDDip. The results of the branch-based student models for OKDDip and OKDDip+KD are reported at the first row and the results of the network-based student models for OKDDip and OKDDip+KD are reported at the second row.

\begin{table}[htbp]
	\caption{Classification error rates (Top-1, \%) of ensemble predictions with network-based student models on CIFAR-10.}
	\label{NetworkE-C10}
	\centering
	\resizebox{0.98\columnwidth}{!}{
		\begin{tabular}{ccccc}
			\toprule
			Network & Baseline & DML & OKDDip \\
			\midrule
			DenseNet-40-12  & 5.54 $\pm$ 0.02 & 5.83 $\pm$ 0.01  &
			\textbf{5.30 $\pm$ 0.07} \\
			ResNet-32  		& 5.15 $\pm$ 0.08 & 6.41 $\pm$ 0.11  &
			\textbf{4.94 $\pm$ 0.05} \\
			VGG-16 			& 5.14 $\pm$ 0.09 & 4.99 $\pm$ 0.03  & 
			\textbf{4.92 $\pm$ 0.02} \\
			ResNet-110  	& 4.39 $\pm$ 0.06 & 4.82 $\pm$ 0.05  & 
			\textbf{3.99 $\pm$ 0.02} \\
			WRN-20-8  		& 4.76 $\pm$ 0.02 & 5.00 $\pm$ 0.05  & 
			\textbf{4.67 $\pm$ 0.03} \\
			\bottomrule
	\end{tabular}}
\end{table}

\begin{table}[htbp]
	\caption{Classification error rates (Top-1, \%) of ensemble predictions with network-based student models on CIFAR-100.}
	\label{NetworkE-C100}
	\centering
	\resizebox{0.98\columnwidth}{!}{
		\begin{tabular}{ccccc}
			\toprule
			Network & Baseline & DML & OKDDip \\
			\midrule
			DenseNet-40-12  & 24.67 $\pm$ 0.13 & 25.70 $\pm$ 0.01  &
			\textbf{23.74 $\pm$ 0.12} \\
			ResNet-32  		& 28.69 $\pm$ 0.15 & 26.47 $\pm$ 0.26  &
			\textbf{22.77 $\pm$ 0.01} \\
			VGG-16 			& 23.31 $\pm$ 0.02 & \textbf{21.62 $\pm$ 0.16}  & 
			22.33 $\pm$ 0.02 \\
			ResNet-110  	& 20.57 $\pm$ 0.09 & 20.81 $\pm$ 0.15  & 
			\textbf{19.55 $\pm$ 0.01} \\
			WRN-20-8  		& 20.47 $\pm$ 0.15 & 19.29 $\pm$ 0.06  & 
			\textbf{19.63 $\pm$ 0.07} \\
			\bottomrule
	\end{tabular}}
\end{table}

\begin{table}[htbp]
	\caption{Classification error rates (Top-1, \%) of ensemble predictions with branch-based student models on CIFAR-10.}
	\label{BranchE-C10}
	\centering
	\resizebox{0.98\columnwidth}{!}{
		\begin{tabular}{cccc|c}
			\toprule
			Network & CL-ILR  & ONE  & OKDDip & Ind\\
			\midrule
			DenseNet-40-12	
			& 7.15 $\pm$ 0.11 & 6.99 $\pm$ 0.05 & \textbf{6.65 $\pm$ 0.14} & 7.15 $\pm$ 0.15 \\
			ResNet-32 	 	
			& 5.9 $\pm$ 0.05 & 5.75 $\pm$ 0.16 & \textbf{5.32 $\pm$ 0.07} & 5.42 $\pm$ 0.25  \\
			VGG-16       	
			& 6.23	$\pm$ 0.10 & 6.15 $\pm$ 0.08 & \textbf{5.91 $\pm$ 0.03} & 6.04 $\pm$ 0.01 \\
			ResNet-110 
			& 4.73 $\pm$ 0.09 & 4.83 $\pm$ 0.08 & 4.33 $\pm$ 0.04 & \textbf{4.30 $\pm$ 0.04} \\
			WRN-20-8 & 5.33	$\pm$ 0.12 & 5.27 $\pm$ 0.03 & 5.18 $\pm$ 0.11 & \textbf{5.21 $\pm$ 0.01} \\
			\bottomrule
	\end{tabular}}
\end{table}

\begin{table}[htbp]
	\caption{Classification error rates (Top-1, \%) of ensemble predictions with an additional teacher.}
	\label{KDE-Table}
	\centering
	\begin{tabular}{ccc}
		\toprule
		DataSet 	& CIFAR-10 & CIFAR-100 \\
		\midrule
		Baseline  	& 5.15 $\pm$ 0.08 & 28.69 $\pm$ 0.15 \\
		KD			& 5.98 $\pm$ 0.15 & 26.35 $\pm$ 0.19 \\
		\midrule
		\multirow{2}{*}{OKDDip} 
		& 5.42 $\pm$ 0.25 & 23.45 $\pm$ 0.23 \\
		& \textbf{4.94 $\pm$ 0.05} & \textbf{22.77 $\pm$ 0.01} \\
		\midrule
		\multirow{2}{*}{OKDDip+KD} 
		& 5.04 $\pm$ 0.03 & 22.92 $\pm$ 0.10 \\
		& \textbf{4.85 $\pm$ 0.03} & \textbf{22.49 $\pm$ 0.09} \\
		\midrule
		Teacher 	& 4.36 $\pm$ 0.02 & 20.72 $\pm$ 0.13 \\
		\bottomrule
	\end{tabular}
\end{table}
\subsection{Diversity Comparison}

Figure \ref{fig:diversity} plots the results of peer diversity comparison with four approaches for WRN-20-8 on CIFAR-10 and CIFAR-100. The diversity of ``Ind'' can be regarded as an upper bound since the student models are trained independently.

\begin{figure}
	\centering
	\subcaptionbox{CIFAR-10\label{fig:diversity_32}}{
		\includegraphics[width=0.48\linewidth]{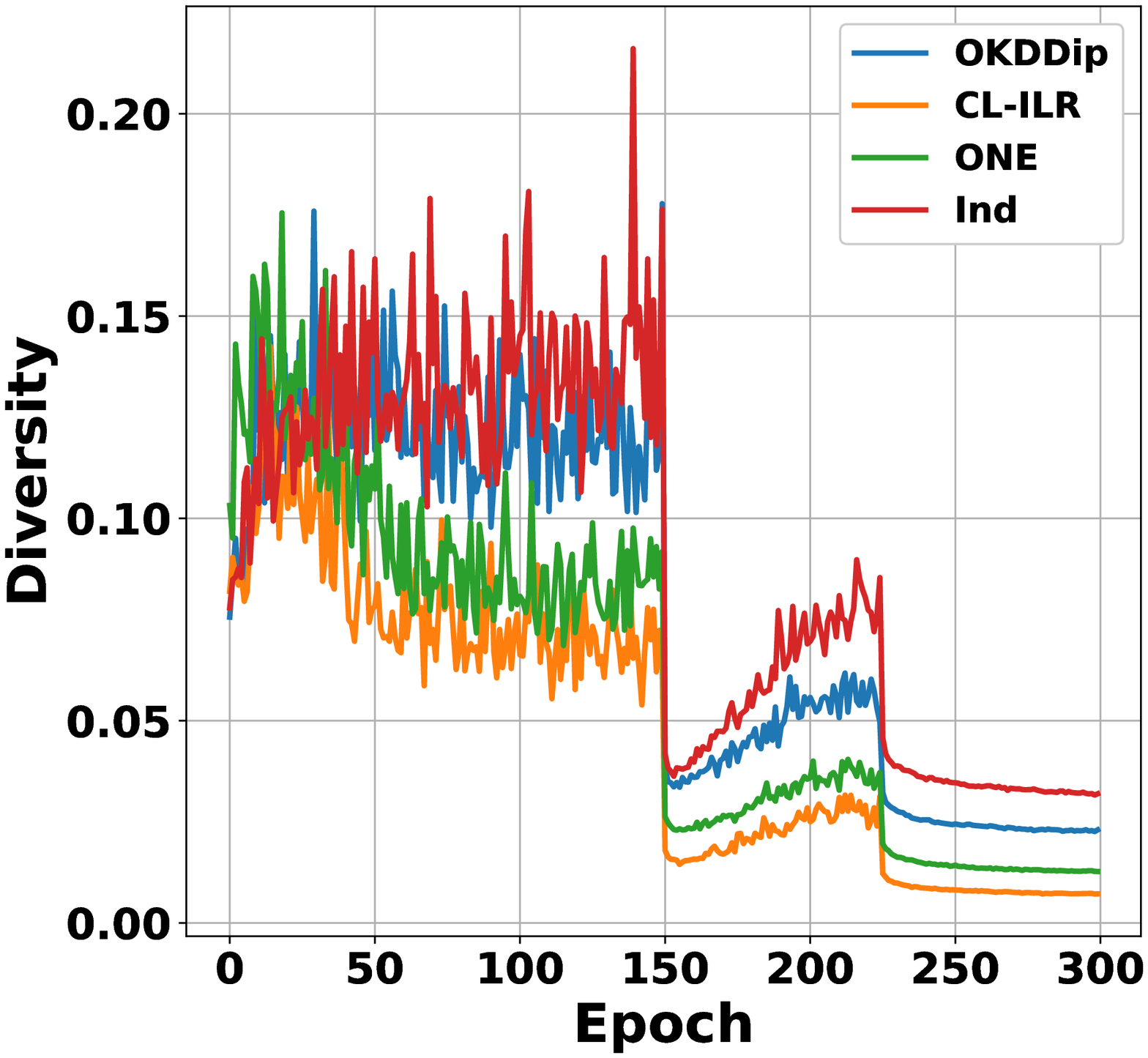}}
	%\hspace{0.1\textwidth}
	\subcaptionbox{CIFAR-100\label{fig:diversity_110}}{
		\includegraphics[width=0.48\linewidth]{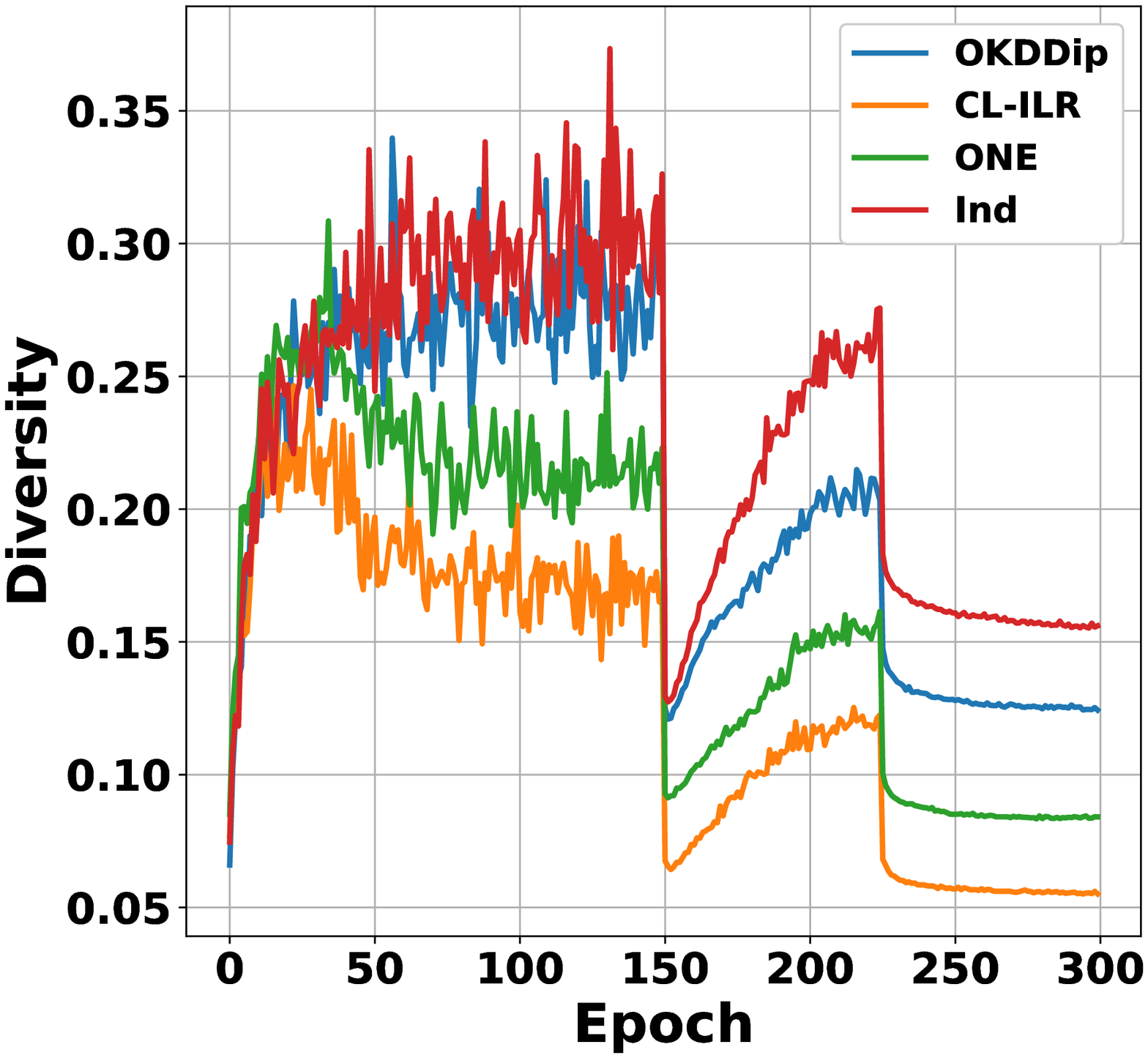}}
	\caption{Peer diversity comparison with branch-based models during training for WRN-20-8.}
	\label{fig:diversity}
\end{figure}

\end{document}